# A method for the ethical analysis of brain-inspired AI

Farisco, M.[1,2], Baldassarre, G.[3], Cartoni, E.[3], Leach, A.[4], Petrovici, M.A.[5], Rosemann, A.[4], Salles, A.[1,6], Stahl, B.[7,4], van Albada, S. J.[8,9]

[1] Centre for Research Ethics and Bioethics, Uppsala University, Uppsala, Sweden

[2] Biogem, Biology and Molecular Genetics Research Institute, Ariano Irpino (AV), Italy

[3] Laboratory of Computational Embodied Neuroscience, Institute of Cognitive Sciences and Technologies, National Research Council of Italy, Rome, Italy

[4] Centre for Computing and Social Responsibility, De Montfort University, Leicester, UK

[5] Department of Physiology, University of Bern, Bern, Switzerland

[6] Institute of Neuroethics

[7] School of Computer Science, University of Nottingham, UK

[8] Institute of Neuroscience and Medicine (INM-6) Computational and Systems Neuroscience & Institute for Advanced Simulation (IAS-6) Theoretical Neuroscience & JARA-Institut Brain Structure-Function Relationships (INM-10), Jülich Research Centre, Jülich, Germany

[9] Institute of Zoology, University of Cologne, Cologne, Germany

*Corresponding Author: Michele Farisco, michele.farisco@crb.uu.se*

**Keywords**: Brain-inspired AI; NeuroAI; AI Ethics; Philosophy of AI; Neuromorphic Computing

## Abstract

Despite its successes, to date Artificial Intelligence (AI) is still characterized by a number of shortcomings with regards to different application domains and goals. These limitations are arguably both conceptual (e.g., related to underlying theoretical models, such as symbolic vs. connectionist), and operational (e.g., related to robustness and ability to generalize). Biologically inspired AI, and more specifically brain-inspired AI, promises to provide further biological aspects beyond those that are already traditionally included in AI, making it possible to assess and possibly overcome some of its present shortcomings. This article examines some conceptual, technical, and ethical issues raised by the development and use of brain-inspired AI. Against this background, the paper asks whether there is anything ethically unique about brain-inspired AI.
The aim of the paper is to introduce a method that has a heuristic nature and that can be applied to identify and address the ethical issues arising from brain-inspired AI. The conclusion resulting from the application of this method is that, compared to traditional AI, brain-inspired AI raises new foundational ethical issues and some new practical ethical issues, and exacerbates some of the issues raised by traditional AI.

**Introduction**

Because intelligence is a biological phenomenon, Artificial Intelligence (AI) is biologically inspired in its very essence. Nevertheless, AI may mimic biology to varying degrees. In some cases, AI research might take direct and extensive inspiration from biology, and from neuroscience in particular. In other cases it might attempt to go beyond biological instantiations of intelligence to implement mechanisms and types of intelligence not present in biological systems. Still, the mutual relationship between neuroscience and AI has been important for advancing both approaches (Hassabis, Kumaran, Summerfield, & Botvinick, 2017; Macpherson et al., 2021; Summerfield, 2023) and has been recently indicated as crucial for the next-generation AI in particular (Zador et al., 2023).

At the outset of AI research in the early 1950s, "the only known systems carrying out complex computations were biological nervous systems" (Ullman, 2019). Accordingly, AI researchers productively used knowledge about the brain as a source of inspiration when seeking to create intelligent systems. This is true especially for AI paradigms alternative to the symbolic approach (also named "Good Old-Fashioned AI" or GOFAI), which was prevalent at the beginning of AI research and aimed to reproduce the logical aspects of intelligence at a high functional level while neglecting the underlying brain mechanisms. In particular, research on artificial neural networks (ANNs) took inspiration from the mechanisms of brain functioning, such as the possibility of processing information based on multiple simple units similar to neurons, and their all-or-none signalling (Kleene, 1956; McCulloch & Pitts, 1943). At the same time, brain-inspired AI continued to inject new ways of thinking about how the brain works, in particular about neural computation (Saxe, Nelli, & Summerfield, 2020), suggesting new mechanisms for neural network models, alongside new research strategies in neuroscience.

This pattern has repeated multiple times, for example also leading to the proposal of Deep Neural Networks (DNNs), which have architectures inspired by the hierarchical structure of perceptual cortices, and currently yield state-of-the-art performance in several Machine Learning (ML) areas. Even though most principles underlying DNNs are still grounded in early neural network models, today we are witnessing an explosion of their potential and applications, mainly owing to the increased availability of data ("Big Data") and to the possibility of using specialised hardware such as Graphics Processing Units (GPUs). Many researchers believe that DNNs offer new possibilities for a mutual collaboration between neuroscience and AI insofar as they learn through sensory signals that are processed in ways that have some resemblance to sensory processing in the brain. This is true especially, but not exclusively, for unsupervised learning and reinforcement learning (Hassabis et al., 2017).

The increased collaboration between neuroscience and AI could result in further improvement of AI, possibly allowing it to overcome some of its current limitations. As mentioned above, the reverse is also possible, with AI inspiring a deeper understanding of the brain (B. A. Richards et al., 2019). Indeed, brains have advantages with respect to current AI models such as a greater capacity for generalisation and for learning from few examples (George, Lazaro-Gredilla, & Guntupalli, 2020), and a much lower power consumption (Attwell & Laughlin, 2001).

Biological inspiration may come from organisms, processes, and phenomena occurring at different spatial and temporal scales, in addition to human brains and cognitive reasoning (Floreano & Mattiussi, 2008). As Floreano and Mattiussi claim, relevant sources of information are evolutionary, cellular, neural, developmental, immune, behavioral, and collective systems, among others.

Notwithstanding their great potential, biologically inspired AI in general, and brain-inspired AI in particular, are not immune to criticism. As the history of science and technology shows, the usefulness of deriving inspiration from biology cannot be taken for granted (Crick, 1989; "Is the brain a good model for machine intelligence?," 2012), and it is theoretically possible for AI to develop along lines not consistent with brain architecture and functioning. Indeed, AI algorithms and techniques are often developed with the aim of solving particular tasks and only later are they compared with the brain (George et al., 2020; Gershman, 2023; Lillicrap, Santoro, Marris, Akerman, & Hinton, 2020).

Calling for AI to emulate the brain thus risks being too reductive and limiting depending on the goals pursued, as AI may fruitfully follow directions different from biology. Taking the brain as the privileged source of inspiration for further development of AI may present problems such as ending up in replicating the brain's limitations and biasing the development of AI. Conversely, assuming that AI could inform a better understanding of the brain raises issues as well. For instance, some authors note that the relevance of DNNs for neuroscience is limited due to the characteristics of current approaches which lack sufficiently constrained learning rules, regularization principles, or architectural features (Hasson, Nastase, & Goldstein, 2020; B. A. Richards et al., 2019; Saxe et al., 2020). Current DNN strategies are thus limited in inspiring empirically testable hypotheses for brain research.

This article reflects upon the question of which conceptual, technical, and ethical issues arise during the development and use of biologically inspired AI, focusing on the specific case of brain-inspired AI. The analysis results from the unique combination of different disciplines characterizing the EU-funded Human Brain Project, which has been inspired by an inter- and multi-disciplinary approach. The research question underlying the ethical analysis is: is there anything ethically unique about brain-inspired AI? The paper starts with a discussion about biologically inspired and brain-inspired AI. Next, it introduces a method for the analysis of the practical ethical issues arising from AI in general and brain-inspired AI in particular. This is followed by an illustration of the method through two case studies (natural language processing and continual learning/context understanding). The article closes with an analysis of the fundamental ethical issues arising from brain-inspired AI, with two main *foci*: concepts and goals.

The conclusion is that, compared to traditional AI, brain-inspired AI can raise qualitatively new foundational ethical issues and some new practical ethical issues, and exacerbates some of the issues raised by traditional AI.

We propose the combination of two ethical approaches (i.e., fundamental and practical) in order to identify emerging ethical issues, prioritize their assessment, anticipate their impact on society, and eventually maximize the benefits derived from brain-inspired AI.

**Biologically inspired and brain-inspired AI: definition and philosophical reflections**

*Definition of biologically and brain-inspired AI*

A preliminary reflection on the meaning of biologically inspired AI and of brain-inspired AI is required before discussing their technical feasibility and related ethical dimensions.

In its broadest sense, biological inspiration refers to the compatibility of AI with current knowledge in biology, particularly in neurobiology. Such a general description, while useful for the sake of introducing the concept, is not sufficiently constrained to be technically operationalizable. More specifically, an AI system is biologically inspired when its architecture and functioning include biological constraints that make specific parts of the system biologically realistic. Importantly, a biologically inspired AI system does not necessarily fully emulate or replicate the reference biological system, since different levels of biological realism are possible. Even if in theory, as mentioned above, biological inspiration can come from many different biological systems (Floreano & Mattiussi, 2008), the main trend today is to define biological realism of AI with specific reference to known biological principles of the brain, in particular mammalian and human brains. Of course, there is no such thing as *the* brain as brains vary substantially between both species and individuals of the same species. Furthermore, different organizational levels and regions of the same brain have different properties.

The differences between brains have various causes. Roughly speaking, brain differences between different species arise from evolutionary factors. Genetic and epigenetic factors (Changeux, Courrège, & Danchin, 1973; Changeux, Goulas, & Hilgetag, 2021), further developmental factors (Bondurianksy & Day, 2018), and factors reflecting interaction with the environment like learning, nutrition, and disease, explain the brain differences between individuals of the same species. Moreover, the brain is a complex organ with a multilevel organization, including molecular, cellular, microcircuit, macrocircuit, and behavioral levels (Amunts et al., 2019). It is likely that not all these levels are equally relevant to the development of brain-inspired AI. This implies three things in particular: (1) any general operational principle of the brain that we identify is the result of a statistical analysis or selection; (2) selection means choice, and this raises the issue of the standards used to make the choice; (3) it is necessary to clarify what level of the brain is taken as the reference for brain-inspired AI.

Biological inspiration conceived as a set of target features from the brain needs to be contextualized and specified. To illustrate: biological inspiration can be achieved by emulating a particular biological mechanism such as the spike-timing-dependent plasticity observed in biological synapses (G. Bi & Poo, 2001; G. Q. Bi & Poo, 1998; Gerstner, Kempter, Van Hemmen, & Wagner, 1996; Schemmel, Grubl, Meier, & Mueller, 2006; S. Song, Miller, & Abbott, 2000), or by using a biologically constrained model, such as the Hodgkin-Huxley model of neuronal action potentials(Hodgkin & Huxley, 1952).

It is also likely that more precise and technically relevant definitions of biological inspiration depend on the specific AI technique considered, and that there is a continuum of possible technologies with different levels of biological inspiration. These different levels are arguably characterized by specific limitations and may raise different types of issues, including ethical issues.

As mentioned above, the biological brain, and more specifically the human brain, is usually uncritically assumed as the standard reference, either explicitly or implicitly. This tendency is manifested, for instance, in these words by Jeff Hawkins: "From the moment I became interested in studying the brain, I felt that we would have to understand how it works before we could create intelligent machines. This seemed obvious to me, as the brain is the only thing that we know of that is intelligent"(Hawkins, 2021). Yet, what seems evident to Hawkins and many others is actually not uncontested, both because it is not obvious that biological inspiration should be limited to the (human) brain (Floreano & Mattiussi, 2008) and because it is not obvious that biological inspiration should be taken as a necessary requirement for improving AI. The fact that it is not obvious that AI should be inspired by the brain can be illustrated in different ways. For instance, spikes themselves might not be necessary in silico, because copper makes for a better signal transmission substrate than cell membranes. Also, post-synaptic potentials (PSPs) do not have to decay gradually, because rectangular PSPs might lead to better functional performance of a network, but are simply more difficult to achieve with the electrophysiology of biological synapses.

### *Differences between current AI and the human brain*

When compared to the human brain, current AI reveals a number of differences and limitations with regards to different domains and goals (Zador et al., 2023). These limitations are arguably of two main kinds: technical and conceptual. The technical limitations depend on the current technological stage of AI and are likely to be reduced and possibly overcome through further progress of knowledge and emerging technology. The conceptual limitations depend on the AI paradigms used, so overcoming them may require revised or new paradigms. To illustrate this, present AI is still narrow, that is, it works for specific tasks in particular domains for which it is programmed and trained, and fails if environmental conditions are different from the training context (Marcus & Davis, 2019). In this respect, its impressive success in specific applications is not yet translated into the capacity for solving broader and more general tasks. For some AI systems, this limitation might not be intrinsic but rather due to the time needed to adapt to new conditions and to learn how to behave in new environments, analogously to what happens with humans. In any case, as a consequence of such constraints, even if omnipresent in our day-to-day lives, the applicability of AI in the real world and its related impact are still limited. Likewise, AI reliability has further development potential, for both technical and conceptual reasons. Indeed, some researchers argue for new approaches and paradigm changes in current AI (Marcus &

Davis, 2019), outlining that one of the main challenges for AI is building learning machines that are as *flexible and robust* as the biological brain and that have the same capacity for *generalization* (Sinz, Pitkow, Reimer, Bethge, & Tolias, 2019). Moreover, in contrast to AI, which at present can interact with a limited subset of variables within a contextualized environment, biological systems respond to a wide variety of stimuli over long periods of time, and their responses alter the environment and subsequent responses, giving rise to a kind of *systemic action-reaction cycle* ("Is the brain a good model for machine intelligence?," 2012).

Other relevant abilities of the human brain that current AI systems are not able to fully replicate are "*multi-tasking*, *learning with minimal supervision,* [...] all accomplished with high efficiency and *low energy cost*", as well as the ability to *communicate* via natural language (Poo, 2018).

One possible strategy to improve the performance of current AI along the dimensions considered above is "to introduce structural and operational principles of the brain into the design of computing algorithms and devices" (Poo, 2018). Relevant results have been obtained through neuromorphic systems, some of which show important advantages including increased computational power per unit of energy consumed and robust learning (Buhler et al., 2017; Cramer et al., 2022; Esser et al., 2016; Frenkel & Indiveri, 2022; Göltz, Kriener, Sabado, & Petrovici, 2021; J. Park, Lee, & Jeon, 2019; Renner, Sheldon, Zlotnik, Tao, & Sornborger, 2021; Wunderlich et al., 2019). Yet the implementation of this strategy is not simple, and the choice itself of taking biology, and specifically the brain, as a reference standard to calibrate and further develop AI systems needs a sound justification that goes beyond the acknowledgment that brains are the only thinking objects we are aware of. In fact this does not *a priori* exclude the possibility of an AI regulated by different principles. Also, bio-inspired AI itself may outperform biology because it can profit from "better" (for example, faster) hardware (Billaudelle et al., 2021; Göltz et al., 2021; Kungl, Schmitt, et al., 2019).

***The risk of putting too much emphasis on emulating the brain***

Focusing on general AI, some researchers have recently argued that biological inspiration and neuroscientific constraints should be regarded as strict requirements for AI until we understand the nature of intelligence (Hole & Ahmad, 2021)). In other words, since we do not yet understand the nature of intelligence, we should take inspiration from the main example of intelligence we have in nature, that is, from the brain. This argument seems to put too much emphasis on the need to define or explain natural intelligence mechanisms before producing AI. This point was already criticized by Turing, who famously elaborated his imitation game to show that the priority is to operationalize intelligence rather than to provide a theoretical definition of it (Dietrich, Fields, Sullins, Van Heuveln, & Zebrowski, 2021; Turing, 1950).

Philosophically, postulating that biological inspiration and resemblance to the human brain in particular should be paradigmatic for AI suggests an implicit endorsement of a form of anthropocentrism and anthropomorphism, which are both evidence of our intellectual self-centeredness and of our limitation in thinking beyond what we are (or what we think we are).

The debate around the biological plausibility of backpropagation is illustrative of the possibility of achieving results similar to biological intelligence by using principles that are not fully biologically compatible. Despite the fact that exact backpropagation is unlikely to happen in the brain (Crick, 1989)(but see (Haider et al., 2021; Lillicrap, Cownden, Tweed, & Akerman, 2016; Lillicrap et al., 2020; Millidge, Tschantz, & Buckley, 2022; Payeur, Guerguiev, Zenke, Richards, & Naud, 2021; Pozzi, Sander, & Roelfsema, 2020; Sacramento, Costa, Bengio, & Senn, 2018; G. Song, Xu, & Lafferty, 2021) for a more informed discussion on this point), AI systems using backpropagation have achieved results comparable to and even better than humans in specific tasks (Pozzi et al., 2020).

From an operational point of view, the question of how to measure biological inspiration also arises. It seems that the quantification of biological inspiration depends on the domain and the brain level taken into consideration, so an AI system can be more or less biologically inspired depending on the particular level of its architecture to which we refer, as well as on the particular aspect of brain architecture and dynamics/physiology that inspired the AI system. To illustrate this, it is possible to assess the biological realism of an ANN by checking if and how it abstracts from the behavior of biological neurons. While this strategy might be sufficient to conclude that ANNs are biologically realistic from an operational point of view, it does not exclude that ANNs are not biologically realistic at another level, for instance at the computational level, since it is possible to implement different computations with the same underlying neuronal behavior. Similarly, it is possible to produce the same network activity with different circuitry (Prinz, Bucher, & Marder, 2004), the latter point being very relevant for neuromorphic systems (Petrovici et al., 2014).

**A method for the ethics of brain-inspired AI**

For the sake of our analysis, ethics can be understood as the attempt to systematize and justify concepts of right and wrong, and to clarify the implications of these concepts for behavior. We pursue this general goal through the combination of a reflection on fundamental/foundational and practical issues: the discrimination between right and wrong relies on the identification of relevant values and principles, which make it possible to justify such distinction. As part of this justification, the definition of key notions like moral subject, moral reasoning, and moral action, among others, plays a crucial role. This definitional task is central in *fundamental* (or *foundational*) *ethics* dealing with the foundation, nature, and evolution of moral thought and judgment, which can be distinguished from *practical ethics* dealing with concrete issues (e.g., ethical assessment of particular fields)(Evers, 2007).

Therefore, we can broadly distinguish two main kinds of ethical issues arising from AI, including brain-inspired AI: foundational (i.e., concerning both the justification of brain-inspired AI and its impact on how we think about fundamental moral notions) and practical (i.e., concerning the impact of brain-inspired AI on our daily life). The first kind of issues involves theoretical analysis, while the second kind of issues involves applied analysis, that is, the use of ethical theory to identify and address the practical issues related to the use of AI. Below, we begin by introducing a method for the analysis of the practical issues raised by AI and show how it can be applied by focusing on two case studies. We then present a foundational ethical analysis of key concepts and goals underlying brain-inspired AI.

*Practical ethics of brain-inspired AI*
Generally speaking, the practical issues raised by a technology can either be intrinsic to the technology itself (i.e., inseparable from how it works) or extrinsic (i.e., emerging from its use in different contexts, from the resulting relationship that humans have with it, from how the technology impacts society at a number of levels, like economic, psychological, environmental, etc., and from their intended goals). We propose that the practical issues arising from biologically inspired AI, including brain-inspired AI, can be organized in terms of (at least) the following main levels:
- *Operational*, related to how AI works;
- *Instrumental*, related to how people use AI;
- *Relational*, related to how people see AI and to the resulting psychological and metaphysical human-AI relationship[1];
- *Societal*, related to the social and economic costs and consequences of the development and use of AI.

Importantly, the distinction between these levels is mainly for the sake of analysis, while in practice the same factor may appear on different levels, for instance in terms of its "proximal" and "distal" effects. To illustrate, brain-inspired AI may have the proximal positive effect of needing less energy which may lead to the distal effect of cheaper systems and a consequent more democratic access to it.

Also, the identified levels should not be seen as confined to biologically plausible technologies. The discussion of ethical aspects of AI can be traced back to the beginning of digital and potentially autonomous systems (Wiener, 1954). It has developed alongside technical progress (Dreyfus, 1972; Whitby, 1991) and gained prominence in recent years when the spectacular successes of machine learning became clear. There are various ways of approaching the ethics of AI (Coeckelbergh, 2020; Dignum, 2019; Stahl, 2021) which include the distinction between foundational and practical and the four levels proposed here.

Operational issues of AI are those that have their basis in the very nature of AI. The focus of established discussions of AI ethics are current machine learning technologies. These are characterized by the need for large datasets for training and validating models. Where such datasets contain personal information, an operational issue would be the intrinsic threat of data protection violations (EDPS, 2020; Kaplan & Haenlein, 2019). Another operational issue arises from the nature of neural networks used in much machine learning, which are complex, opaque, and often not open to scrutiny, leading to a large discourse on explainability of AI (Friedrich, Mason, & Malone, 2022; Yeung, 2018).

Instrumental concerns may have their roots in these operational questions and touch on how AI is used. The most prominent example of this is probably the problem of biases and discrimination. Bias in machine learning can result from the replication of bias in social relationships that is reflected in the data and hence in machine learning models. When such models are used unthinkingly, this may lead to the replication of bias, for example, in job selection or law enforcement applications (Birhane, 2021; Team, 2018). Societal issues typically arise less from the nature of AI and more from the way they influence and shape existing socio-economic structures. Relevant concerns can refer to economic injustice and unfair distribution of risks and benefits (Walton & Nayak, 2021; Zuboff, 2019). Other types of issues arise when AI is used to inappropriately influence democratic processes or contribute to power concentration (Nemitz, 2018; Parliament, 2020). Further examples include the environmental impact (Nishant, Kennedy, & Corbett, 2020) of AI or its potential to change the future of warfare (Brundage et al., 2018; L. Richards, Brockmann, & Boulanini, 2020).

---
[1] The way people perceive AI has an immediate impact on their own self-understanding and also on the way people conceive their nature in relation to the nature of AI.

Relational issues where AI changes the way we as humans see ourselves or our relationship with other humans, technology, or the world at large also have a prominent history. This includes speculative discussions about the potential of future AI, which are linked to concepts such as transhumanism, singularity, or superintelligence (Bostrom, 2014). On a more immediate level, AI can serve as a metaphor for thinking about humans and changes to the capabilities of AI then change what characteristics we ascribe to ourselves and to each other.

We now focus on how these issues are raised in the context of brain-inspired AI.

**Table 1. Practical ethical issues potentially arising from brain-inspired AI**

| OPERATIONAL LEVEL | |
|---|---|
| **POTENTIAL BENEFITS** | **POTENTIAL RISKS** |
| - *More environmentally friendly systems*, because brain-inspired AI may requires less power than more traditional AI<br><br>- *More rapid optimization of systems* that are more strongly constrained than traditional AI (i.e., with a more restricted space of solutions for optimization) | - *Limited capacity of brain-inspired AI to emulate the brain*, leading to the risk of its insufficient recognition and flawed communication about it<br>- *Limited possibility for humans to optimize brain-inspired AI*, because it may be more strongly constrained than traditional AI and there is the risk of *a priori* excluding factors that are crucial for achieving the desired goals |
| INSTRUMENTAL LEVEL | |

| POTENTIAL BENEFITS | POTENTIAL RISKS |
|---|---|
| Applicability to certain domains for which traditional AI is less suited | *Risk of new kinds of brain-based crimes*, because brain-inspired AI may significantly improve the technology-mediated understanding of brain features and eventually lead to their exploitation (e.g., for brain hacking) |

<div align="center">**RELATIONAL LEVEL**</div>

| POTENTIAL BENEFITS | POTENTIAL RISKS |
|---|---|
| *Possibility to inspire more positive attitudes towards AI*, because users' awareness of the 'brain-inspired' nature of the AI systems they use may raise more empathetic feelings | - *Risk of hyped perception and misplaced trust in brain-inspired AI*, because of the risk of anthropomorphic attitudes and because of the background view of the brain as a paradigm of efficiency and effectiveness combined with inadequate recognition of and communication about the limited capacity of brain-inspired AI to emulate the brain<br>- *Bewildering impact on ingrained beliefs about human identity*, particularly about human exceptionalism |

<div align="center">**SOCIETAL LEVEL**</div>

| POTENTIAL BENEFITS | POTENTIAL RISKS |
|---|---|
| *Possibility of new, cheaper, and more democratic products*, including commercial applications, because brain-inspired AI may need less energy and | *Risk of increased concentration of power in a few hands*, because of both the advanced technology required by brain-inspired AI and its application domain |

| may develop a broad domain of application | extending that of more traditional AI |
|---|---|

Brain-inspired AI arguably promises new benefits and raises new risks for society (Doya, Ema, Kitano, Sakagami, & Russell, 2022). In particular, there are some practical ethical issues raised by traditional AI that might be exacerbated by brain-inspired AI, while others seem more specific to it. For example, intrinsic limitations and shortcomings of the brain might lead to specific practical ethical issues if somehow replicated by brain-inspired AI, as described in Table 1.

At the operational level, brain-inspired AI presents a number of potential benefits and risks. Among the possible benefits, brain-inspired AI has the potential to lead to *more environmentally friendly* systems because it may require less power than more traditional AI. Also, there is the possibility of *more rapid optimization of systems* that are more strongly constrained than traditional AI (i.e., with a more restricted space of solutions for optimization).

On the other hand, brain-inspired AI has a limited capacity to emulate the brain because it necessarily refers to a specific level of brain organization (e.g., cellular level) or to specific phenomena and mechanisms, ignoring or streamlining its relation to other lower or higher levels (e.g., molecular or synaptic level) or to other relevant phenomena and mechanisms. Also, brain activity is not independent from bodily influences (e.g., bodily activity seems to influence conscious perception (H.-D. Park & Tallon-Baudry, 2014) and confidence in subjective perception (Allen et al., 2016)), and the molecular environment and metabolism regulate these brain-body couplings (Haydon & Carmignoto, 2006; Jha & Morrison, 2018; Petit & Magistretti, 2016). This operational limitation raises the risk for developers and the general public of *insufficient recognition of and flawed communication about the possibility of failure in fully emulating the brain*. The fact that brain-inspired AI possibly fails to fully emulate the brain has ethical implications because there is a tendency, especially in lay people, to conceive the brain as paradigmatic (i.e., as a model of efficiency and effectiveness). It is possible that this view is projected onto brain-inspired AI if its above-mentioned limitation is not adequately recognized and communicated to the public. In other words, an ethical problem arising from the limited capacity of brain-inspired AI to emulate the brain is the lack of explicit and fair communication about this limitation. The tendency to assign human-like trustworthiness to Large Language Models (LLMs) like ChatGPT is an illustration of this risk.

The necessary selection of specific levels of brain organization as reference/target for brain-inspired AI can also *limit the possibility for humans to optimize the system if it fails to achieve its goals*: such failure might be caused by the fact that achieving the goal in question depends on factors (i.e., brain features) we are blind to because they were excluded by the preliminary selection (e.g., AI might be emulating one level without including its relation with other levels). In other words, if relevant aspects of brain organization or dynamics are excluded *a priori*, and they are crucial for optimizing a specific function, then there is limited possibility for developers to optimize the system (i.e., to improve its capacity to achieve the desired goal).

For example, it has been observed that it is useful for synapses to have access to multiple variables, beyond mere pre- and postsynaptic activity. One such type of variable is the activity of dendritic compartments, which are represented in morphologically complex neuron models, but not in point neuron models. Thus, choosing the simpler (point) model might impair the ability of a network to learn hierarchical representations (Haider et al., 2021; Sacramento et al., 2018). Furthermore, morphologically complex neurons can implement entire multilayer artificial neural networks (Beniaguev, Segev, & London, 2021; Häusser & Mel, 2003; Poirazi, Brannon, & Mel, 2003).

Another illustration of the point above is the inclusion of short-term plasticity or cortical oscillations into the dynamics of a modeled stochastic network. Without these, as the network learns to solve increasingly complex problems, it can become overly confident in one solution and is impaired in its ability to consider alternative solutions (Korcsak-Gorzo et al., 2022; Leng & Kakadiaris, 2018).

A third example concerns the learning of time series, which relies on the presence of long enough transients in a network, often implemented by neuronal adaptation. If all specific time constraints are shorter than the transients in the signals to be learned (i.e., if all memory mechanisms are too short), learning becomes difficult or impossible (Bellec et al., 2020; Maass, Natschlager, & Markram, 2002).

At the instrumental level, brain-inspired AI promises *applicability to certain domains for which traditional AI is less suited* because of the lack of necessary features, such as a sufficient capacity for online problem-solving and generalizability (i.e., the capacity to learn on the fly and to adapt to contextual variables that are different from the training data).

On the other hand, the features of AI that could make its deployment more widespread and beneficial in some respects (like online problem solving) may have undesirable consequences if used for negative purposes, like fraudulent applications. In addition to the risks raised by traditional AI (e.g., security reduction, exploitation/violence, misuse/dual use, reduction or loss of human agency), *brain-inspired AI may raise additional risks of brain-based crimes*, because brain-inspired AI may significantly improve the technology-mediated understanding of brain features and eventually lead to their exploitation (e.g., for brain hacking, that is unauthorized access to and influence on subjective mental states/operations).

At the relational level, on the one hand, since it is modeled on the (human) brain, brain-inspired AI can inspire more empathetic feelings and this can lead to *more positive attitudes towards AI* (i.e., people are more willing to use AI) because of anthropomorphic tendencies. Considering that some cultures are still characterized by a form of technophobia(Weil & Rosen, 1995), *awareness that AI is brain-inspired might diminish existing technophobic attitudes*.

On the other hand, the anthropomorphic feelings raised by brain-inspired AI may cause a *misplaced trust in its capabilities as well as hypes about its application*, because of the background view of the brain as a paradigm of efficiency and effectiveness combined with inadequate recognition of and communication about the limited capacity of brain-inspired AI to emulate the brain. Furthermore, brain-inspired AI might be perceived as so close to human identity so as to have a *bewildering impact on ingrained beliefs about the relationships between humans and machines.* For instance, brain-inspired AI may be perceived as further evidence that human exceptionalism is illusory. This risk might again create either *disproportionate expectations* or *fears* about brain-inspired AI.

At the societal level, because of its potentially great energy efficiency and an impressive ability to handle vast amounts of data (Mehonic & Kenyon, 2022), brain-inspired AI has the *potential for leading to new as well as to cheaper systems and commercial products*, which may eventually result in *more democratic and participative processes*. For instance, brain-inspired AI may lead to new clinical tools accessible to an increased number of stakeholders because of reduced costs compared to analogous traditional AI solutions. Yet the level of scientific knowledge and the kind of technology necessary to develop brain-inspired AI are so advanced compared to the technology available today that they raise the *risk of increased concentration of power in a few hands*. The development of brain-inspired AI requires significant resources of different kinds, including economic and human resources. Adequate financial investments, both private and public, are necessary, and not all countries can afford such expenses, and some prefer to privilege other lines of research (Mehonic & Kenyon, 2022). Furthermore, brain-inspired AI is cross-, multi-, and interdisciplinary, and it requires the collaboration of researchers from different fields. Not all public research institutions have the possibility to set up the necessary collaborative groups, and it is possible that only economically strong private companies can afford the related costs. Similarly, the potentially new application domains of brain-inspired AI or its improved performance in existing domains may amplify the uneven distribution of power between the rich and the poor that comes with powerful technology in general.

This analysis of the practical ethical issues arising from brain-inspired AI with reference to four main levels can be illustrated through the consideration of two possible fields of application that are benefitting from brain-inspired advancements: natural language processing and continual learning/context understanding.

*Natural language processing*

LLMs like GPT-3 and GPT-4 (Brown et al., 2020) have recently gained a lot of popularity because of impressive progress in their applicability, even by lay people in ordinary life contexts. For instance, the discussion about ChatGPT has exploded in the last few months. After its release at the end of 2022, a growing number of people have used it for a variety of purposes, including assistance in academic research (e.g., for summarizing or writing texts), in medicine (e.g., for assistance in diagnosis), in business (e.g., for content marketing), in education (e.g., for reviewing students essays), and in computer science (e.g., for coding). Notwithstanding these impressive results, ChatGPT is still brittle and fragile. This is caused by the fact that the technology is still in development as well as by the fact that it relies only on one type of training data (i.e., text data). Compared to human intelligence, ChatGPT lacks embodiment, that is the sensorimotor strategies that humans use in order to explore the world through multisensory integration within a constitutive brain-body-environment interaction (Pennartz, 2009). As a result of this interaction, human intelligence is multidimensional, and it is capable of online learning: developing a realistic representation of the world that is adapted in real time. ChatGPT and LLMs in general presently lack this multidimensional and multisensory representation of the world, and therefore they are intrinsically exposed to a limited and distorted knowledge.

Brain-inspired solutions might assist the improvement of LLMs on these aspects. To name just a couple of examples, potentially relevant results have been obtained incorporating biologically inspired neural dynamics into deep learning using a novel construct called spiking neural unit (SNU) (Woźniak, Pantazi, Bohnstingl, & Eleftheriou, 2020). SNU has allowed improving the energy efficiency of AI hardware, and it promises improvements in a number of tasks, including natural language processing. Also, a novel online learning algorithm for deep Spiking Neural Networks (SNNs), called online spatio-temporal learning (OSTL), has been introduced, with potential for improving language modeling and speech recognition (Bohnstingl, Wozniak, Pantazi, & Eleftheriou, 2022).

The application of the ethical model described above to the case of natural language processing leads to the identification of the following ethically relevant potential benefits and risks (see Table 2).

At the *operational level*, an increased ability of AI systems to process natural language is likely to result in increased efficiency due to the use of heuristics based on frequency of occurrence in natural language. On the other hand, there is the potential risk of greater difficulty of testing systems because there is a larger set of possible commands for which to check the functionality, leading to biased or inappropriate output going unnoticed.

At the *instrumental level*, there is the possibility of an easier use of AI (e.g., a richer and more flexible vocabulary to operate it) and more possibilities to exploit it (e.g., greater ease of use in varied contexts). In fact, AI systems will likely interact with human users in a more intuitive and direct way because more instructions/training data mediated by natural language will be possible.

At the same time, it is also likely to increase the risk of unethical data processing and handling, for instance through more invasive AI systems, that is, systems characterized by an increased ability to identify and process sensitive data from humans.

At the *relational level*, an increased ability of AI systems to process natural language may either result in leading to more positive attitudes towards them (if they are perceived as closer to humans) or to hyped perception and misplaced trust in AI capacities (if anthropomorphic biases and unbalanced communication about the actual capacity of AI systems prevail). Also, an increased risk of the so-called "uncanny valley" arises: a sense of uneasiness or eeriness experienced by humans when a technology is very similar to a human being but something appears "off" (Ciechanowski, Przegalinska, Magnuski, & Gloor, 2019).

Finally, at the *societal level*, optimized natural language processing is likely to result in more accessible and "democratic" AI systems (i.e., more people will be able to use them because the interaction will be much easier and possibly less expensive because of an increased commercialization), but at the same time it might trigger new forms of economic exploitation (e.g., through unequal capacity for user profiling) and a powerful impact on the job market (e.g., raising the risk of increasingly replacing humans in more "creative" activities like journalism or coding).

**Table 2. Potential ethically relevant benefits and risks arising from improved natural language processing by brain-inspired AI**

|  BENEFITS | RISKS |
|---|---|
| **OPERATIONAL LEVEL** | |
| Increased efficiency due to the use of heuristics based on frequency of occurrence in natural language | Greater difficulty of testing systems because there is a larger set of possible commands for which to check the functionality, leading to biased or inappropriate output going unnoticed |
| **INSTRUMENTAL LEVEL** | |
| Easier use of AI (e.g., a richer and more flexible vocabulary to operate it) and more possibilities to exploit it | Increased risk of unethical data processing (e.g., privacy infringement) |
| **RELATIONAL LEVEL** | |
| More positive attitudes towards AI systems | Risk of hyped perception and misplaced trust in AI systems<br><br>Increased risk of "uncanny valley" |
| **SOCIETAL LEVEL** | |
| More accessible AI systems (both easy to use and less expensive) | New forms of economic exploitations |

*Continual learning and context understanding*
Continual learning (also named open-ended learning) and context understanding present significant challenges to current AI, which is limited in its ability to learn new things on the fly and to adapt to new circumstances. Recent research has highlighted the possible contribution of neuroscience to improving the learning capacity of AI systems. For instance, the flexibility, adaptiveness to new circumstances, and fast learning capacity characteristic of the brain might derive in part from its capacity to autonomously learn on the basis of "intrinsic motivations" (e.g., curiosity, interest in novel stimuli or surprising events, and interest in learning new behaviors) (Barto, 2004; Santucci, Baldassarre, & Mirolli, 2013). These are different from "extrinsic motivations" involving biological drives, such as hunger and pain, directed toward obtaining specific resources from the environment. Intrinsic motivations are maximally evident in children at play, and are, for example, related to novelty, surprise, and the success in accomplishing desired goals. The biological function of intrinsic motivations is to drive the learning of knowledge and skills that might be later used to find useful resources. The digital simulation of intrinsic motivations allows AI systems, such as DNNs and humanoid robots, to actively seek the knowledge they lack. This lets them adapt to new circumstances without the need for external guidance, eventually making AI more autonomous. Moreover, it speeds up learning processes as these are directed to the interactions that maximize knowledge and competence gain.

In addition to the inspiration provided by the brain's intrinsic motivations for improving AI learning capability, recent research considers the biological feature that we are not born as clean slates, but already have a brain structure optimized by evolution for learning from our experiences. For instance, innate aspects of neural network structure may already gear networks toward effective task performance (Stöckl, Lang, & Maass, 2022), and allow the generalization of learning within families of tasks (Bellec et al., 2020). Furthermore, cortical network models have been used to infer different possible learning rules (Zappacosta, Mannella, Mirolli, & Baldassarre, 2018), including in spiking networks (Jordan, Schmidt, Senn, & Petrovici, 2021). The spiking feature would add biological plausibility to most current AI systems, as would the inclusion of layer-specific connectivity as found in the cerebral cortex, including in particular layer-specific interactions between feedforward and feedback signals across cortical areas (Markov et al., 2014; Rao & Ballard, 1999).

Directly related to the improvement of learning capability is the AI capacity for context understanding, that is, its ability to know relevant features of real-world contexts in order to appropriately act in them and to interact with other systems/agents within them. Time-continuous learning in substrates with time-continuous dynamics is the subject of a number of recent studies (Haider et al., 2021; Kungl, Dold, Riedler, Senn, & Petrovici, 2019; Sacramento et al., 2018; Senn et al., 2023). This obviates the need for phases in learning, allowing AI agents to simply observe their surroundings in a time-continuous fashion, making it easier to embed them in complex, real-time contexts, which is one of the aspects current AI is missing.

Research has shown the advantages of attention for limiting the information stream to be processed, which may aid continual learning and context understanding. This is not only evidenced by the aforementioned LLMs relying on so-called transformers, which differentially weight different parts of their input, but also by direct investigation of human learning (Niv et al., 2015). In this vein, a bio-inspired model that integrates bottom-up and top-down attention to scan the scene has been introduced (Ognibene & Baldassare, 2015), showing the advantages of developing an integrated interplay between the two by continual reinforcement learning. This can lead an agent to autonomously find relevant stimuli in order to solve tasks, thus processing a smaller amount of information and speeding up learning. Another example is the recently proposed Attention-Gated Brain Propagation (BrainProp), a form of reinforcement learning that obviates the need for supervision to approximate error backpropagation (Pozzi et al., 2020). Attentional mechanisms are likely to become even more relevant as AI systems are developed that handle multimodal input streams, including not only text but also for instance auditory and visual input.

The application of the ethical model presented above results in the following ethically relevant potential issues (see Table 3).

At the *operational level* the capacity for continual learning and the improvement of context understanding will likely result in more autonomous, flexible, and adaptable AI systems, which will be less prone to operative failures and more user-friendly. This will likely be counterbalanced by an increased risk of less transparent AI systems: since they will be more able to learn on the fly how to act, they will have evolving parameters and will eventually be more independent from top-down instructions and from external monitoring, reducing the space for human understanding and supervision. This is an ethical issue because increased opacity of AI systems leads to an increased risk of unexpected operational failure with potential negative consequences, and fewer possibilities to prevent them.

At the *instrumental level*, the potential uses of AI (i.e., the contexts in which it can be used) will eventually be increased, but there is the risk that the capacity for controlling it by the users will be reduced.

At the *relational level*, an enhanced ability of AI for continual learning and for interacting with its surroundings might promote on the one hand a general perception of AI as more reliable and robust because it would be more able to adapt to changing external conditions. On the other hand, some people might feel more insecure, because of an increasing perception of AI systems as autonomous agents, and a self-perception as less able than AI in an increasing number of activities.

At the *societal level*, optimized continual learning and context understanding will make it possible to maximize the advantage of using AI in many more contexts than traditional AI, but the risk of replacing human agents in different sectors, including those requiring more creativity and capacity for adaptation, will likely increase, eventually leading to concerns about a reduction or loss of human agency.

**Table 3. Potential ethically relevant benefits and risks arising from improved continual learning and context understanding by brain-inspired AI**

| BENEFITS | RISKS |
|---|---|
| **OPERATIONAL LEVEL** ||

| More autonomous, flexible, adaptable AI systems | Less transparent AI systems |
|---|---|
| **INSTRUMENTAL LEVEL** ||
| Increased number of contexts in which to use AI systems | Less control on AI systems by the users |
| **RELATIONAL LEVEL** ||
| People more prone to see AI systems as reliable tools | Increased feeling of being insecure and/or less able than AI systems |
| **SOCIO-ECONOMIC LEVEL** ||
| Possibility to maximize the advantage of using AI in different contexts | Risk of replacing human agents also in more creative activities |

**Fundamental ethics of brain-inspired AI**

We have introduced four levels for the identification and the analysis of the practical ethical issues arising from brain-inspired AI, and have illustrated them with a couple of case studies. Here we identify some fundamental/foundational ethical issues raised by brain-inspired AI. As mentioned above, these issues concern the justification of the attempt itself to build brain-inspired AI and its impact on how we think about fundamental moral notions. Therefore, we distinguish two main categories of foundational ethical issues:

- **Those related to goals**, which refer to questions like: What is the driver of brain-inspired AI? What do we want to achieve by it?

- **Those related to concepts**, which include issues like epistemic risk, implicit assumptions about brain-inspired AI, and considerations about the historical, cultural, and societal contexts of brain-inspired AI.

*Fundamental ethical issues emerging from brain-inspired AI in relation to goals*

Different goals motivate the attempt to translate brain principles and features into AI, and each raises fundamental ethical issues (see Table 4).

**Table 4. Illustrative goals of brain-inspired AI and related fundamental ethical issues.**

| Goal of brain-inspired AI | Fundamental ethical issue |
| --- | --- |
| Improving traditional AI | What is the assumed meaning of improvement? E.g., does it include awareness of the potential impact on different social groups or actors, including non-human? |
| Making AI more autonomous | How might more autonomous AI systems impact human autonomy, including how people think about their autonomy? |
| Discovering operational principles in different sensorimotor and cognitive fields, which may be engineered and applied to AI systems | Does brain-inspired AI aim to replicate operational brain principles related to ethical reasoning? Can this introduce artificial moral agents? |
| Taking the brain as a model for advancing in the direction of Artificial General Intelligence (AGI) | How should AGI be conceptualized, particularly in relation to ethical dimensions? |

As noted above, a first, main intention of the development of brain-inspired AI is *improving traditional AI*, by advancing in the following sectors (among others):

- More effective interaction of AI systems with the world
- Optimized information processing
- Improvement of AI systems' capability of on-line problem solving
- Improvement of high assurance systems, like in manufacturing, information technology, navigation, etc.
- More effective real-world applications, like in banking, defense, education, finance, medicine, security, etc.
- More efficient robotics applications and better embodiment of AI
- More flexible and autonomous AI

A fundamental issue arising from the general goal of improving traditional AI concerns the *underlying understanding of improvement*. Related to this question is the issue of which improvements will lead to a better society. These are conceptual issues with direct ethical implications. For instance, to what extent does the desired improvement include considerations about potential societal and ethical impacts of brain-inspired AI on different social groups or actors, including non-human? Is the notion of improvement in this case only informed by technical considerations?

A second goal of brain-inspired AI is to *make AI more autonomous* through a better clarification and artificial replication of motivations, emotions, and values underlying human decision-making. A fundamental ethical question raised by this goal revolves around *possible impacts of more autonomous AI on humans*, including on how humans perceive their autonomy.

A third goal of brain-inspired AI is more knowledge-oriented: *discovering operational principles in different sensorimotor and cognitive fields, which may be engineered and applied to AI systems*. In this way a positive epistemic feedback loop between brain-inspired AI and neuroscience may be realized: the first is inspired by the second and at the same time contributes to further advancing our knowledge of the brain. A fundamental ethical question concerns the *kind of brain principles that we aim to scale and apply to AI systems*: do they also include principles related to the capacity for ethical reasoning? If so, the possibility that brain-inspired AI systems will be equipped with moral reasoning cannot be logically excluded, even if the technical feasibility remains challenging.

Finally, another motivation for brain-inspired AI is *advancing in the direction of Artificial General Intelligence (AGI)*, even if both its conceptual reliability and its technical feasibility remain controversial (Summerfield, 2023). A fundamental point to consider is that general intelligence is likely shaped by several factors, including bodily and environmental factors. This implies that limiting the focus to the brain risks being overly reductive and eventually ineffective for reaching AGI. At the ethical level, the question of *what kind of generality we refer to in the underlying concept of AGI and why we are seeking it* arises: does it also include ethically relevant features and if so, how are these dealt with? And what is the underlying motivation of seeking AGI?

The aforementioned goals are not morally neutral in themselves. If achieved, they will necessarily lead to ethically relevant discussions regarding, for example, the notion of autonomy in general and of AI autonomy in particular, and about the necessity to align human and AI values. While this type of debate is not new within AI ethics, brain-inspired AI has the potential to bring it to the fore and exacerbate it insofar as it promotes advances in the direction of autonomous and ethically reasoning AI.

Importantly, all these goals, even if intrinsically bearing on ethical issues, are *a priori* neutral with regards to potential positive or negative societal consequences. In fact, the technical improvements potentially deriving from brain-inspired AI can equally lead to both good or bad applications. It is crucial to facilitate and to implement a multidisciplinary reflection and to set up a monitoring system in order to anticipate and identify relevant ethical issues in a timely manner. We propose the combination of the two ethical approaches introduced in this paper (i.e., fundamental and practical) as a promising strategy in order to identify emerging ethical issues, prioritize them, anticipate their impact on society, and eventually maximize the benefits deriving from brain-inspired AI.

*Fundamental ethical issues emerging from brain-inspired AI in relation to concepts*

**Table 5. Foundational ethical issues arising from brain-inspired AI**

| **Fundamental issues** | **Related ethical questions** |
|---|---|
| The brain as a model for AI | How can we be sure that applying brain principles in AI is the best possible option for advancing the development of AI and then maximizing the benefit for our society? |
| | Are we maybe just seconding our anthropocentric bias which makes it hard if not impossible for us to think beyond the form of intelligence as we know it in nature? |
| | How unlikely is it that we will eventually end up replicating the shortcomings that make biological, and particularly human intelligence a source of risks and dangers for human society and the rest of the world? |
| Potential implications of brain-inspired AI for those concepts that are traditionally assumed as the basis for qualifying an agent as moral | If what characterizes and qualifies our biological intelligence can be realistically replicated, i.e. simulated, artificially, would it imply that there is nothing unique in our identity? |
| | Would our self-understanding as moral agents be impacted? |
| | If what characterizes us as moral agents (e.g., relevant cognitive and/or emotional abilities) can be artificially replicated through a brain-inspired AI, would that imply that this is also a moral agent? |
| | Would a hypothetical artificial replication of our moral agency have consequences for the way to discriminate between what is good and |

|  | what is bad? |
|---|---|

This second group of fundamental ethical issues refers to theoretical considerations, which include epistemic risk, implicit assumptions about brain-inspired AI, and considerations about the historical, cultural, and societal contexts of brain-inspired AI. Epistemic risk means the possibility that the target model of brain-inspired AI (i.e., the brain) is not the best reference to optimize AI, either because we do not sufficiently know it or because of its own intrinsic limitations. This connects to implicit assumptions about brain-inspired AI, more specifically to the possibility to take for granted that the brain is exemplary and that taking inspiration from it is the best possible strategy for optimizing AI. Finally, a number of questions arise from the interaction of brain-inspired AI with particular historical, cultural, and social contexts, and the resulting impact on ingrained foundational moral notions.

A first fundamental question concerns the justification of brain-inspired AI: why take the biological brain as a model for AI in the first place? This question is not only scientifically and technically relevant, but it is also important from the point of view of public policy, of Science and Technology Studies (STS), and of ethics. From a public policy perspective a key question is whether the expected benefits coming from brain-inspired AI are sufficient to prioritize its development rather than focusing resources on the further development of traditional AI. From an STS perspective, a point to address is what are the underlying politics and narratives used to frame brain-inspired AI as promising or cutting-edge. From an ethical perspective, the issue becomes whether biologically plausible and brain-inspired AI will actually result in the maximization of the benefit for society. After all, we cannot rule out the possibility that by developing brain-inspired AI we might be just following our anthropocentric bias which makes it hard if not impossible for us to think beyond the form of intelligence as we know it in nature. If so, we cannot exclude that we tend to perceive brain-inspired AI as the best strategy for the benefit of society not because this is actually the case, but because of our inability to see and appreciate alternatives to biological intelligence. In other words, we may have a tendency to consider brain-inspired AI as more promising than traditional AI not for technical reasons but for cultural biases. Moreover, we must be aware of a possibility that has practical implications but also an important theoretical dimension: brain-inspired AI might eventually end up replicating the shortcomings that make biological, and particularly human intelligence, a source of risks and harms for human societies at large. These ethical questions touch upon our view of ourselves and our nature as intelligent agents, as well as our view of what is the best for society and what role science and technology should play in it.

A second fundamental question concerns the possible implications of brain-inspired AI for those concepts that are traditionally assumed as the basis of morality, namely the conditions for agency in general and for moral agency in particular. Intentionality (i.e., wilful goal-oriented action) is usually assumed as a necessary and sufficient condition for agency (Davidson, 1963, 1971). Yet this traditional view of agency has been complemented and even criticized from different perspectives, highlighting, for instance, that the ability to reflect on and to care about the motivations of an action are peculiar to persons vs. non-persons (Frankfurt, 1971), and that agency comprises different dimensions (i.e., authenticity, privacy, responsibility, trust) that are distinguished and reciprocally linked at the same time (Schönau et al., 2021). Attributing this plethora of concepts to AI is not uncontroversial, and brain-inspired AI might push us further in this direction, for instance if it replicates relevant functions of the human brain.

With specific regards to moral agency, brain-inspired AI might raise issues concerning both the traditional criteria for qualifying as a moral agent, and the sense and/or self-concept of being a moral agent. Moral agency is basically understood as the ability to discern right from wrong and to choose the relevant goals to pursue. Whilst the issue of whether this capacity is mainly cognitive or emotional continues to be debated, there is wide consensus on the crucial role played by the brain and particularly by some specific cerebral areas and related functionalities (Verplaetse, 2013).

A number of ethically relevant foundational questions may arise from the prospect of brain-inspired AI: if what characterizes and qualifies our biological intelligence can be realistically replicated (i.e., simulated) artificially, would this imply that there is nothing unique (both metaphysically and morally) to our natural identity? The supposed uniqueness of humans has already been falsified by evolutionary biology, but brain-inspired AI poses a different threat, specifically to the supposedly unique identity of humans as moral agents. Specifically, if what characterizes us as moral agents (e.g., relevant cognitive and/or emotional abilities) can be artificially replicated through a brain-inspired AI, would that imply that AIs can be moral agents? Would our self-understanding as moral agents be impacted? And would a hypothetical artificial replication of our moral agency have consequences for the way we discriminate between good and bad?

It is unlikely that implementing some brain-inspired principles in currently narrow AI will raise these kinds of issues, particularly because at present AI appears too constrained to the specific goals for which it has been pre-programmed, and its flexibility and robustness are too limited to enable the rise of artificial moral agency. These issues are more likely to be raised by the presence of a hypothetical human-like or human-level brain-inspired AI which at least at present appears far-fetched. Still, the theoretical possibility is, as such, ethically relevant, because it might be translated into technical feasibility, and ethics should not be taken to be just reactive but, importantly, proactive, anticipating possible future scenarios.

There is an additional fundamental issue potentially relevant to brain-inspired AI, one that arises from how brain-inspired AI is conceived and developed, and which necessarily impacts how it eventually operates. Since genetic, epigenetic, and environmental factors, including culture, socialization, and culturally/historically formed notions of identity shape the brain, the possibility of introducing biases in brain-inspired AI arises. This raises a number of ethically relevant issues: can brain-inspired AI inadvertently present a set of novel biases, that is, different from those raised by traditional AI? If so, how to evaluate them ethically? For instance, brain-inspired AI may enhance biases caused by the selection of training data (e.g., it may take only selected neuronal data as reference) if calibrated on particular sets of brain data (e.g., from adult brains rather than young brains, without accounting for important differences in terms of plasticity and dynamics).

The problem raised by brain-inspired AI biases can also be turned around, when raising the question of whether AI will be able to understand and apply the necessary forms of "diversity" and the "situatedness" of human cognition and perception in various types of AI-based predictions and applications. The point, which is ethically very salient, is that without awareness or understanding of the significance of the diversity of human knowledge, perception, and capacity of making sense of the world, many AI predictions and applications are themselves likely to be biased, of limited utility or eventually useless.

Also, either implicit or explicit neuro-essentialist and neuro-reductionist cultural models (Vidal, 2017) might impact how the general public perceives brain-inspired AI. In fact, if the brain is conceived as the core essence of human nature, and if human identity is eventually reduced to the brain, then imitating it means imitating what humans are, including ethically relevant features. As a consequence, brain-inspired AI might raise new fears, for instance about the supposed risk of creating artificial sentience leading to artificial forms of suffering (Metzinger, 2021).

**Conclusion**

Brain-inspired AI is an attractive strategy for improving on current AI, but it raises a number of technical and ethical issues. In this paper we summarized the main conceptual and technical aspects of brain-inspired AI, and then we provided a method for its ethical analysis, distinguishing two main kinds of ethical issues: fundamental/foundational and practical. Brain-inspired AI has the potential to raise new fundamental/foundational and practical ethical issues, as well as to exacerbate the practical ethical issues raised by traditional AI, which should be considered in relation to this promising approach.

**Acknowledgments**

European Union's Horizon 2020 Framework Programme for Research and Innovation under the Specific Grant Agreement No. 945539 (Human Brain Project SGA3) (All); European Innovation Council Action No.101071178 (CAVAA) (M.F.); European Union Grant Agreements No. 604102, 720270, 785907 (M.A.P.); the Manfred Stärk Foundation (M.A.P.).